\documentclass[letterpaper, 10 pt, conference]{ieeeconf}  

\IEEEoverridecommandlockouts                              

\usepackage{times}

\usepackage{cite}

\usepackage{xspace}
\usepackage{multicol}
\usepackage[bookmarks=true]{hyperref}
\usepackage{amssymb}
\usepackage{xcolor}
\usepackage{graphicx}
\usepackage[caption=false,font=footnotesize]{subfig}
\usepackage{amsmath, amssymb, amsfonts, mathtools}

\usepackage{amsthm}

\newcommand{\ipp}{IPP\xspace}
\newcommand{\cipp}{C-IPP\xspace}
\newcommand{\lipp}{LIPP\xspace}

\allowdisplaybreaks

\theoremstyle{plain}
   
\theoremstyle{remark}

\theoremstyle{definition}

\theoremstyle{plain}

\begin{document}


\title{LIPP: Load-Aware Informative Path Planning with Physical Sampling}

\author{\authorblockN{Hojune Kim}
\authorblockA{School of Electrical and\\Computer Engineering\\
University of Southern California\\
Los Angeles, California \\
Email: hojuneki@usc.edu}
\and
\authorblockN{Guangyao Shi}
\authorblockA{School of Computer Science\\
University of Southern California\\
Los Angeles, California \\
Email: shig@usc.edu}
\and
\authorblockN{Gaurav S. Sukhatme}
\authorblockA{School of Computer Science\\
University of Southern California\\
Los Angeles, California \\
Email: gaurav@usc.edu}}

\maketitle

\begin{abstract}
In classical Informative Path Planning (\cipp), robots are typically modeled as mobile sensors that acquire digital measurements such as images or radiation levels. In this model---since making a measurement leaves the robot's physical state unchanged---the cost of traversing an edge remains static regardless of when it is traversed. This is a natural assumption for many missions, but does not extend to settings involving physical sample collection, where each collected sample adds mass and increases the energy cost of all subsequent motion. As a result, \ipp formulations that ignore this coupling between information gain and load-dependent traversal cost can produce plans that are distance-efficient but energy-suboptimal, collecting fewer samples and less data than the energy budget would permit. In this paper, we first introduce Load-aware Informative Path Planning (\lipp), a strict generalization of \cipp that explicitly models this coupling, with \cipp recovered as the special case of zero sample mass. We then formulate \lipp as a Mixed-Integer Quadratic Program (MIQP) that jointly optimizes visitation location, order, and per-location sampling count under an energy budget. We further derive theoretical bounds on the path-length increase of \lipp relative to \cipp, characterizing the trade-off for improved energy efficiency. Finally, through extensive simulations across 2{,}000 diverse mission scenarios, we demonstrate that \lipp progressively achieves higher uncertainty reduction per unit energy as sample mass increases.
\end{abstract}

\IEEEpeerreviewmaketitle

\section{Introduction}

Robotic exploration missions increasingly rely on autonomous systems to gather information where human access is limited or impossible, such as planetary surface exploration \cite{10783051}, environmental monitoring \cite{Das}, and precision agriculture \cite{van_Essen_2025}. In these settings, robots must continually decide where to move and what data to collect under strict resource limits. Informative Path Planning (\ipp) addresses this by finding paths that maximize information gain under travel-cost constraints.

In classical \ipp\ (\cipp) formulations, sensing is modeled as a digital operation: measurements such as images, temperature readings, or signal density are acquired without altering the robot's physical state. Under this assumption, the amount of data collected is decoupled from the budget constraint, since traversing an edge while carrying many samples costs the same as traversing it while carrying none. As a result, energy and distance are not distinguished, and minimizing path length often serves as a reasonable surrogate for minimizing energy consumption. However, this abstraction breaks down in missions involving physical sample acquisition. Consider a lunar rover collecting regolith samples for laboratory analysis. Each additional sample increases the rover's payload, raising the energy cost of all subsequent motion. The same effect arises for surface vehicles accumulating water samples or drones retrieving agricultural specimens for field inspection. In each of these settings, information is acquired in the form of physical samples rather than purely digital observations, and each collected sample modifies the robot's physical state in a way that persists over the remainder of the mission.

\begin{figure}[t]
    \centering
    \includegraphics[width=\columnwidth]{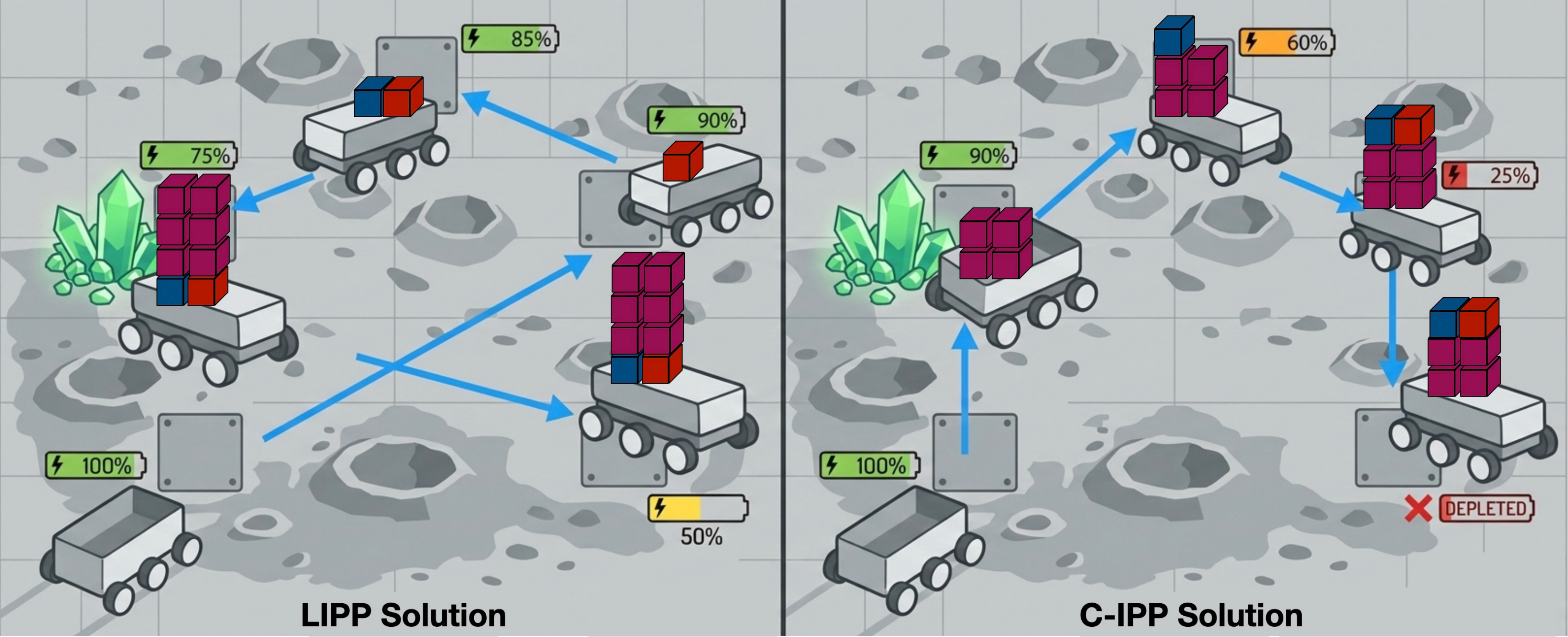}
\caption{Illustration of load-aware (\lipp) and load-unaware (\cipp) planning in a regolith sampling mission in which planetary rovers collect and store soil samples. As the rover gathers multiple samples at a scientifically important site (where minerals are concentrated in the figure), the accumulated payload increases the energy required for the rest of the journey. \cipp does not account for this effect during planning, producing distance-efficient but energy-inefficient paths. In contrast, \lipp explicitly models the evolving load and introduces an additional decision dimension: how much material to collect at each visited location. This added flexibility enables \lipp to allocate more samples to high-value regions while maintaining or even reducing the overall energy budget, resulting in greater information gain per unit of energy expended compared to classical distance-based planning.}
\label{fig:example}
\end{figure}

In addition, many real-world missions collect \emph{multiple} samples at scientifically significant locations, adding a further decision dimension: how much to collect at each site. In \cipp\ this choice would be inconsequential, since sampling does not affect traversal cost; under load-dependent energy, however, it fundamentally alters the planning problem, as heavier collection raises the cost of every subsequent edge. Routing, visitation order, and sampling amount thus become tightly intertwined and must be optimized jointly. To this end, we propose Load-aware \ipp\ (\lipp), a strict generalization of \cipp\ that explicitly models the impact of physical sample mass on energy consumption.

The main contributions of this paper are as follows:
\begin{itemize}
    \item We propose \lipp, a new class of \ipp problem in which the amount of physical sampling affects estimation uncertainty and induces load-dependent, order-sensitive traversal costs.
    
    \item We derive an exact MIQP reformulation of \lipp.
    
    \item We derive theoretical bounds quantifying the path-length loss of \lipp\ relative to distance-optimal solutions.
    
    \item We validate \lipp\ across 2{,}000 simulated scenarios, showing up to $3\times$ higher uncertainty reduction per unit energy than \cipp. Our implementation code are publicly available at \url{https://github.com/hojunieee/lipp-load-aware-ipp}.
\end{itemize}

\section{Related Work}
\ipp\ has been widely applied to environmental monitoring, spatial field estimation, and target tracking, where paths are optimized under resource constraints to reduce estimation uncertainty or to accurately recover distributional statistics (e.g., quantiles) of the underlying field \cite{9832756, Hitz2017AdaptiveCI, Rueckin2022AdaptiveIPP, Hollinger2014SamplingBased, JakkalaAOA26, shi2023robust, shek2024drop}. These methods differ chiefly in how the path is optimized. Discrete, optimization-based formulations operate on a graph and select sensing locations under a travel budget, including the sparse-regression mixed-integer program of Dutta et al.\ \cite{10916979} that we build upon, while continuous-space methods optimize waypoints directly by gradient descent, using sparse Gaussian processes to scale to large environments under distance and kinematic budgets \cite{JakkalaA24IPP}. A complementary line addresses the adaptive, online setting, replanning as measurements arrive via deep reinforcement learning \cite{Rueckin2022AdaptiveIPP} or learned attention-based kernels \cite{Chen2024POAM}. Closest to our setting, Ott et al.\ \cite{Ott2023AIPPMS} couple sensing and motion by letting the agent trade sensor accuracy against energy cost under a shared budget. Across all of these, however, measurements are digital: an observation incurs a fixed, action-local cost and leaves the robot's physical state unchanged, so sensing and mobility remain weakly coupled and traversal cost is governed only by path geometry.

Separately, routing problems with load-dependent travel costs have been studied in the operations research literature. Representative examples include pickup-and-delivery problems, vehicle routing with load-dependent fuel consumption, and the Traveling Thief Problem (TTP), where item selection directly influences travel speed or cost \cite{6557681, Bektas2011PollutionRouting, Zachariadis2015LoadDependentVRP}. These formulations explicitly capture the coupling between carried load and traversal effort, typically balancing transportation cost against profit or service objectives. However, in these settings, energy or fuel consumption is treated purely as an operational cost accumulated deterministically along a route, rather than as part of a sequential decision process that trades off information gain and future mobility. Moreover, they generally assume a fully known environment with deterministic rewards, and their objective is cost–profit optimization rather than uncertainty reduction. Consequently, they do not incorporate adaptive information gathering or the estimation-theoretic structure that underpins informative path planning.

Existing \ipp methods do not jointly model repeated physical sampling and cumulative load-dependent traversal energy, while routing formulations with load-dependent travel costs do not incorporate probabilistic field estimation or information-theoretic objectives. Our work bridges these two directions by integrating load-dependent energy consumption into informative path planning for spatial field estimation, yielding a tightly coupled optimization problem that simultaneously determines sampling amount, traversal order, and uncertainty reduction.

\section{Problem Formulation}
\subsection{Problem Setup}
Consider a robot capable of collecting physical samples operating within a closed and bounded region $\mathcal{X} \subset \mathbb{R}^2$.
We discretize $\mathcal{X}$ into a weighted directed graph $G := (V, \mathcal{E}, d)$, where $V$ is a set of $n$ admissible sampling locations, $\mathcal{E} \subseteq V \times V$ is a set of directed edges, and $d : \mathcal{E} \rightarrow \mathbb{R}^{+}$ denotes the traversal cost per unit carried load along each directed edge. The cost function $d$ may incorporate environment-dependent factors such as terrain slope or surface roughness for planetary rovers, as well as wind or water currents for aerial and aquatic robots.

In addition to the sampling vertices $V$, we define a set of $m$ test locations $T \subset \mathcal{X}$ at which we aim to estimate an unknown static scalar field $f : \mathcal{X} \rightarrow \mathbb{R}$. 
We model the field $f$ as a Gaussian Process (GP),
$
f \sim \mathcal{GP}(0, k),
$
where $k : \mathcal{X} \times \mathcal{X} \rightarrow \mathbb{R}$ is a known positive definite covariance kernel, similar to many GP-based \ipp formulations \cite{gp}.

The robot is assigned a start vertex $s \in V$ and a goal vertex $g \in V$. 
The planner produces a path for the robot,
$
P = \bigl\langle (v_{1},\, l_{1}), (v_{2},\, l_{2}), \dots, (v_{p},\, l_{p}) \bigr\rangle,
$
where $v_{1} = s$, $v_{p} = g$, $(v_{j}, v_{j+1}) \in \mathcal{E}$ for all $j = 1,\dots,p-1$, and $l_{j} \in \{0,1,\dots,S_{\max}\}$ denotes the number of unit-mass physical samples collected at vertex $v_{j}$.\\

\noindent\textbf{Problem 1} \textit{Load-Aware \ipp (\lipp)}\\
Given a set of test locations $T$, determine a path $P$ consisting of
$(\text{vertex},\, \text{sample amount})$ tuples that minimizes the posterior
uncertainty of the test locations subject to an accumulated energy budget:
\begin{align*}
\min_{P} \quad & \operatorname{PostVar}_{T}(P) \\
\text{s.t.} \quad & \operatorname{Energy}(P) \le B.
\end{align*}
Here, $B$ denotes the available energy budget, and
$\operatorname{PostVar}_{T}(P)$ represents the trace of the posterior
covariance matrix of the GP evaluated at $T$ after collecting samples
along path $P$.

\subsection{Measurement Noise and Energy Model}
Conventional \ipp formulations assume that the robot acquires a single digital noisy measurement $y_j$ at each visited vertex $v_j$, with constant measurement noise variance:
\begin{equation}
y_j = f(v_j) + \eta_j,
\qquad
\eta_j \sim \mathcal{N}\!\left(0, \sigma^2\right),
\label{eq:measurement_model_1}
\end{equation}
where $\sigma^2$ denotes the variance of the measurement noise associated with each sample.

In realistic sampling settings, where repeated sampling is allowed, collecting multiple independent samples at the same vertex can reduce the effective measurement noise variance due to the variance-reduction effect of averaging independent observations \cite{MoodIntroStat}. To capture this effect, we adopt the following measurement model: if $l_j \in \mathbb{Z}_{\ge 0}$ samples are collected at vertex $v_j$, then the resulting measurement noise is modeled as
$
\eta_j \sim \mathcal{N}\!\left(0, \frac{\sigma^2}{l_j}\right).
$
However, acquiring additional samples also increases the load carried by the robot. 
Consider a path consisting of an ordered sequence of vertices $\{v_1, \dots, v_p\}$, 
where $l_j \in \{0,1,\dots,S_{\max}\}$ denotes the number of unit-mass samples collected at vertex $v_j$. 
Let $\lambda$ denote the unit mass of a sample and $R_0$ denote the mass of the robot without any samples. 
Then, the cumulative mass of the robot at the $i$-th vertex is defined as
$
R_i := \sum_{j=1}^{i} (\lambda l_j)  + R_0.
$
Using this, we define the total energy expenditure along the path as
\[
E :=
\sum_{j=1}^{p-1}
d(v_j, v_{j+1}) \, R_j,
\]
where $d(v_j, v_{j+1})$ denotes the travel cost between consecutive vertices. This formulation provides a first-order approximation of load-dependent energy consumption based on the relation $E = \text{force} \times \text{distance}$, assuming that the dominant traversal forces (e.g., friction and gravity) scale linearly with the robot mass. 

Note that the proposed load-aware formulation strictly generalizes \cipp. 
In particular, when the robot base mass is normalized to $R_0 = 1$ and the sample unit mass approaches zero $\lambda \to 0$, the cumulative mass satisfies $R_j \to R_0 = 1$ for all $j$. 
In this limit, the energy model reduces to $E(P) = \sum_{j=1}^{p-1} d_j$, and the energy budget constraint becomes equivalent to the standard distance budget constraint used in \cipp: $\sum_{e} d(e)\chi_e \le b$. 
Therefore, \cipp can be viewed as a special case of the proposed \lipp formulation in which sampling does not affect mobility.

\subsection{Objective Function}
We quantify information using the posterior variance \cite{10916979} over the test set, 
\begin{equation}
\operatorname{PostVar}_{T}^M(P)
=
\operatorname{trace}\!\left(
M \, \bar{k}_{TT}
\right),     
\label{eq:postvar_1}
\end{equation}
where $M \in \mathbb{R}^{m \times m}$ is a diagonal weight matrix representing the importance of each test vertex. 
The posterior covariance matrix given by kernel is
\begin{equation}
\bar{k}_{TT}
=
k_{TT}
-
k_{TV}
\left(
k_{VV} + N
\right)^{-1}
k_{VT}, 
\end{equation}
where
$
k_{VV} \in \mathbb{R}^{n \times n}, \quad
k_{TV} \in \mathbb{R}^{m \times n}, 
$
and $N \in \mathbb{R}^{n \times n}$ is a diagonal noise matrix whose entries depend on the sampling amount decision $l_j$, and its diagonal element is defined as $N_{jj}=\frac{\sigma^2}{l_j}$ using our new measurement noise model.

However, the objective in \eqref{eq:postvar_1} involves a matrix inverse whose entries depend nonlinearly on the discrete sampling decisions through $N$, resulting in a highly nonlinear and nonconvex problem not directly amenable to standard optimization solvers.

To address this, we leverage the equivalence between the GP posterior mean and the linear least-squares estimator (LLSE) under Gaussian assumptions (Theorem~5 in \cite{10916979}). Crucially, this is not a linear approximation but an exact algebraic identity: under jointly Gaussian assumptions, the LLSE coincides with the conditional expectation, so the reformulation preserves the original objective exactly. The key insight is that the optimal GP predictor at each test location is a linear combination of the observations at visited vertices; collecting this weight matrix into $A\in\mathbb{R}^{m\times n}$, the posterior covariance can be written as a polynomial in $A$ and $N$ without the matrix inverse.

Specifically, we introduce a mixed-integer formulation of Problem 1 as follows.

\begin{align}
\min_{A,N} \quad
& \operatorname{tr}\!\left(
M \bigl(A(k_{VV}+N)A^\top - 2k_{TV}A^\top + k_{TT}\bigr)
\right)
\label{eq:objective}
\\[4pt]
\text{s.t.}\quad
& \sum_{e\in\mathcal E_v^{\mathrm{in}}} \chi_e
=
\sum_{e\in\mathcal E_v^{\mathrm{out}}} \chi_e
\le 1
\label{eq:flow_conservation}
\\
& \sum_{e\in\mathcal E_s^{\mathrm{out}}} \chi_e = 1,
\quad
\sum_{e\in\mathcal E_g^{\mathrm{in}}} \chi_e = 1
\label{eq:start_goal}
\\
& \sum_{e\in\mathcal E_s^{\mathrm{in}}} \chi_e = 0,
\quad
\sum_{e\in\mathcal E_g^{\mathrm{out}}} \chi_e = 0
\label{eq:no_backflow}
\\
& y_v = \sum_{e\in\mathcal E_v^{\mathrm{in}}} \chi_e
\label{eq:vertex_activation}
\\
& -A_{\max} y_v
\le
A_{t,v}
\le
A_{\max} y_v
\label{eq:A_activation}
\\
& o_s = 0
\label{eq:order_start}
\\
& 0 \le o_v \le |V|-1
\label{eq:order_bounds}
\\
& o_v \ge o_u + 1 - \mathcal{M}^{(o)}(1-\chi_{uv})
\label{eq:MTZ}
\\
& L_s = \lambda \sum_{c=1}^{S_{\max}} c\, z_{s,c}
\label{eq:load_start}
\\
& L_v \ge L_u + \lambda
\sum_{c=1}^{S_{\max}} c\,z_{u,c}
- \mathcal{M}^{(L)}(1-\chi_{uv})
\label{eq:load_propagation}
\\
& 0 \le L_v \le L_{\max}
\label{eq:load_bounds}
\\
& R_v = R_0 + L_v
\label{eq:robot_mass}
\\
& l_v = \sum_{c=1}^{S_{\max}} c\,z_{v,c}
\label{eq:load_definition}
\\
& \sum_{c=1}^{S_{\max}} z_{v,c} = y_v
\label{eq:sampling_activation}
\\
& \sum_{(u,v)\in\mathcal E} d_{uv} \,R_u\, \chi_{uv}
\le B
\label{eq:energy_budget}
\\
& \chi_e \in \{0,1\},
\quad e\in\mathcal E
\label{eq:chi_binary}
\\
& y_v \in \{0,1\},
\quad v\in V
\label{eq:y_binary}
\\
& z_{v,c} \in \{0,1\},
\quad v\in V,\; c=1,\dots,S_{\max}
\label{eq:z_binary}
\\
& A \in \mathbb R^{m\times n},\quad
l_v\in\mathbb Z_{\ge0},\quad
L_v\in\mathbb R_{\ge0},\quad
R_v\in\mathbb R_{\ge0}.
\label{eq:variable_domains}
\end{align}
where constraints \eqref{eq:flow_conservation} and 
\eqref{eq:vertex_activation} hold for all $v\in V\setminus\{s,g\}$,
constraint \eqref{eq:A_activation} holds for all $t\in T$ and 
$v\in V\setminus\{s,g\}$, and 
constraints \eqref{eq:MTZ} and \eqref{eq:load_propagation}
hold for all $(u,v)\in\mathcal E$.

In this formulation, $A \in \mathbb{R}^{m\times n}$ is a continuous decision variable representing the linear estimator matrix obtained from the LLSE reformulation. The discrete decision variables are:
(i) binary edge variables $\chi_e$ indicating whether edge $e$ is selected,
(ii) binary vertex variables $y_v$ indicating whether vertex $v$ is visited,
(iii) binary sampling variables $z_{v,c}$ indicating whether $c$ unit samples are collected at vertex $v$,
(iv) integer order variables $o_v$ representing the position of vertex $v$ along the path,
(v) integer sample count variable $l_v$ representing the number of samples taken at vertex $v$,
(vi) continuous load variables $L_v$ representing the mass of the cumulative sample load carried upon departing vertex $v$, and 
(vii) continuous mass variables $R_v$ representing the mass of the robot upon departing vertex $v$.

Constraints~\eqref{eq:flow_conservation}–\eqref{eq:vertex_activation} encode the flow and path feasibility requirements. 
Specifically, \eqref{eq:flow_conservation} enforces flow conservation at intermediate vertices, 
\eqref{eq:start_goal} and \eqref{eq:no_backflow} ensure a single outgoing edge from the start vertex and a single incoming edge to the terminal vertex, and \eqref{eq:vertex_activation} maintains consistency between vertex and edge activation variables. 
Constraint~\eqref{eq:A_activation} links the estimator variables to the visitation decisions via a linear big-$M$ formulation, enforcing $A_{t,v}=0$ whenever vertex $v$ is not selected (i.e., $y_v=0$). Thus, estimator coefficients are active only at visited vertices.

We introduce ordering constraints to eliminate subtours and capture the order-dependent nature of the load–energy coupling induced by repeated physical sampling. 
Constraint~\eqref{eq:order_start} fixes the order of the start vertex to zero, while \eqref{eq:order_bounds} bounds the order variable within $0$ and $|V|-1$. 
Constraint~\eqref{eq:MTZ} enforces precedence along selected edges: if edge $(u,v)$ is active, then the order of vertex $v$ must be at least one greater than that of $u$. 
Here, $\mathcal{M}^{(o)}$ denotes a sufficiently large big-$M$ constant, e.g., $|V|$. 
Together, constraints~\eqref{eq:order_start}–\eqref{eq:MTZ} form a Miller–Tucker–Zemlin (MTZ) subtour elimination scheme \cite{Miller_Tucker_Zemlin_1960}, preventing disconnected cycles and ensuring a single directed path from $s$ to $g$.

We next introduce the load and energy constraints. Constraint~\eqref{eq:load_start} initializes the sample load at the start vertex to the load collected at the first vertex. 
Constraint~\eqref{eq:load_propagation} ensures that the cumulative load at each visited vertex equals the load carried upon arrival plus the samples collected at that vertex.
The Constraint~\eqref{eq:load_bounds} bounds the load variable within its feasible range, and Constraint~\eqref{eq:robot_mass} defines the mass of the robot using the cumulative sample load and its initial mass.
Constraint~\eqref{eq:load_definition} defines the integer sampling amount $l_v$ at vertex $v$ in terms of the binary variables $z_{v,c}$, which is used in the noise matrix $N$ in the objective. 
Constraint~\eqref{eq:sampling_activation} ensures consistency between vertex activation and sampling decisions by enforcing that exactly one sampling level is selected if and only if the vertex is visited. 
Finally, constraint~\eqref{eq:energy_budget} imposes the total energy budget $B$, where the energy expenditure depends on both traversal distance and cumulative robot mass along the path.

With this formulation, we explicitly capture the coupling between uncertainty reduction from repeated sampling and the order-dependent energy cost induced by transporting the accumulated physical load. Nevertheless, the resulting optimization problem remains computationally challenging. The objective function is nonconvex due to the cubic term $A (k_{VV}+N) A^\top$, and the energy budget constraint~\eqref{eq:energy_budget} contains bilinear terms in $R_u$ and $\chi_{uv}$. Consequently, the resulting formulation is a mixed-integer nonconvex program and is NP-hard. To enable an efficient solution using a commercial solver such as Gurobi, we reformulate the problem into an equivalent MIQP: the cubic terms in the objective are reduced to quadratic form via variable disaggregation (auxiliary variables $A_{t,v,c}$), and the bilinear energy constraint is exactly linearized via a McCormick envelope.

\section{Mixed Integer Quadratic Programming Reformulation}

The objective in~\eqref{eq:objective} contains the cubic interaction
$z_{v,c}\,A_{t,v}^{2}$ arising from the product $A\,N\,A^{\top}$,
where the diagonal noise matrix $N$ depends on the discrete sampling
decisions through~$z_{v,c}$.
To reduce this to a quadratic program, we disaggregate each estimator
coefficient into per-sampling-level components by introducing auxiliary
continuous variables
\[
A_{t,v,c}\in\mathbb{R},
\qquad t\in T,\;v\in\mathcal V,\;c=1,\dots,S_{\max},
\]
with the aggregation and big-$M$ linking constraints
\begin{equation}
A_{t,v}
=
\sum_{c=1}^{S_{\max}} A_{t,v,c},
\qquad
-A_{\max}\,z_{v,c}
\;\le\;
A_{t,v,c}
\;\le\;
A_{\max}\,z_{v,c}.
\label{eq:A_aggregation_bigM}
\end{equation}
These constraints ensure $A_{t,v,c}=0$ whenever $z_{v,c}=0$,
linking the estimator coefficients to the discrete sampling choices.

Because both $M$ and $N$ are diagonal, the trace objective
$\operatorname{tr}\!\bigl(M(A(k_{VV}+N)A^{\!\top}
-2\,k_{TV}A^{\!\top}+k_{TT})\bigr)$
decomposes into a sum over test locations~$t$.
Expanding and substituting
$N_{vv}=\sum_{c=1}^{S_{\max}}\frac{\sigma^{2}}{c}\,z_{v,c}$
yields cubic terms of the form $\frac{\sigma^{2}}{c}\,z_{v,c}\,A_{t,v}^{2}$.
By~\eqref{eq:A_aggregation_bigM}, each such term is equivalently
expressed as the quadratic $\frac{\sigma^{2}}{c}\,A_{t,v,c}^{2}$,
since $A_{t,v,c}$ is nonzero only when $z_{v,c}=1$.
The resulting MIQP objective is
\begin{align}
\min_{A_{\cdot,\cdot,\cdot},\,z}\;
\sum_{t=1}^{m} M_{tt}
\Big(\,
&\sum_{v_1,v_2\in\mathcal V}
k_{VV}(v_1,v_2)\,
A_{t,v_1}\,A_{t,v_2}
\nonumber\\
&+\sum_{v\in\mathcal V}
\sum_{c=1}^{S_{\max}}
\frac{\sigma^2}{c}\,A_{t,v,c}^{2}
\nonumber\\
&-2\!\sum_{v\in\mathcal V}
k_{TV}(t,v)\,A_{t,v}
\;+\;
k_{TT}(t,t)
\Big).
\label{eq:MIQP_objective}
\end{align}

\noindent
All constraints~\eqref{eq:flow_conservation}--\eqref{eq:z_binary} carry over
unchanged, except that the bilinear energy budget
constraint~\eqref{eq:energy_budget} is replaced by its exact McCormick
linearization \cite{McCormick_1976}. Introducing auxiliary variables
$w_{uv}\in\mathbb{R}_{\ge0}$ for every $(u,v)\in\mathcal E$, we write
\begin{align}
\sum_{(u,v)\in\mathcal E} d_{uv}\, w_{uv}
&\;\le\; B,
\label{eq:energy_linearized}
\\[1pt]
w_{uv} &\le R_u,
\qquad
w_{uv} \le R_{\max}\,\chi_{uv},
\nonumber\\
w_{uv} &\ge R_u - R_{\max}\,(1-\chi_{uv}),
\qquad
w_{uv} \ge 0,
\label{eq:omega_bounds}
\end{align}
\noindent
where $R_{\max}=R_0+L_{\max}$.
When $\chi_{uv}=1$ the envelope forces $w_{uv}=R_u$; when $\chi_{uv}=0$
it forces $w_{uv}=0$. Since $R_u\in[R_0,R_{\max}]$ is bounded and
$\chi_{uv}$ is binary, the McCormick relaxation is exact, and the
overall formulation remains an MIQP solvable by an off-the-shelf solver
such as Gurobi.

\section{Analysis}
Incorporating repeated physical sampling and load-aware energy modeling fundamentally changes the structure of the \ipp problem, enabling improved energy efficiency while maintaining low posterior variance. These benefits, however, introduce trade-offs, including potentially longer geometric path lengths and increased computational complexity. Although path length no longer directly corresponds to total energy expenditure in the \lipp setting, it remains practically relevant as a proxy for mission execution time. In this section, we provide a theoretical analysis of both trade-offs.

\subsection{Theoretical Bound on Execution Path Length}

We bound the execution path length of \lipp relative to \cipp, where \cipp
denotes \eqref{eq:objective}–\eqref{eq:A_activation} with $N_{jj} = \sigma^2$
and the distance constraint $\sum_{e} d(e)\chi_e \le b$, as in \cite{10916979}.

Let $P_D$ and $P_E$ denote the paths generated by \cipp and \lipp, respectively, with $p_D := |P_D|$ and $p_E := |P_E|$ vertices, and let $S_{\max}$ be the maximum number of unit samples per vertex. We assume that the \lipp solution consumes no more energy than the \cipp solution, i.e., $E(P_E) \le E(P_D)$.
For a path $P = \langle v_1, \dots, v_p \rangle$, define
\[
E(P) := \sum_{j=1}^{p-1} R_j d_j,
\qquad
D(P) := \sum_{j=1}^{p-1} d_j,
\]
where $d_j$ is the travel cost of edge $(v_j, v_{j+1})$ and $R_j$ is the robot
mass while traversing it. Since each visited vertex contributes between $1$
and $S_{\max}$ unit samples of mass $\lambda > 0$ to the base mass $R_0$,
\[
R_0 + \lambda j \;\le\; R_j \;\le\; R_0 + \lambda S_{\max} j,
\]
and hence, for any path $P$,
\begin{equation}
\sum_{j=1}^{p-1} (R_0 + \lambda j)\, d_j
\;\le\; E(P) \;\le\;
\sum_{j=1}^{p-1} (R_0 + \lambda S_{\max} j)\, d_j .
\label{eq:energy_sandwich}
\end{equation}

Since $1 \le j \le p-1$, bounding $j$ termwise in \eqref{eq:energy_sandwich}
and using $E(P_E) \le E(P_D)$ gives
\[
\begin{aligned}
(R_0 + \lambda)\, D(P_E)
&\le E(P_E) \le E(P_D) \\
&\le \big(R_0 + \lambda S_{\max} (p_D - 1)\big) D(P_D).
\end{aligned}
\]
Dividing by $R_0 + \lambda > 0$ yields
\begin{equation}
D(P_E) \;\le\; C\, D(P_D),
\qquad
C := \frac{R_0 + \lambda S_{\max} (p_D - 1)}{R_0 + \lambda},
\label{eq:general_bound}
\end{equation}
which holds for any arbitrary graph structure and vertex selection.

If the graph lives on an equidistant grid, i.e., $d_j = D(P)/(p-1)$ along each path,
then $\sum_{j=1}^{p-1} j\, d_j = \tfrac{p}{2} D(P)$ exactly, and
\eqref{eq:energy_sandwich} sharpens \eqref{eq:general_bound} to
\begin{equation}
\frac{D(P_E)}{D(P_D)}
\;\le\;
\frac{R_0 + \lambda S_{\max} \tfrac{p_D}{2}}{R_0 + \lambda \tfrac{p_E}{2}}.
\label{eq:grid_bound}
\end{equation}

\begin{figure*}[t]
    \centering
    \includegraphics[width=\textwidth]{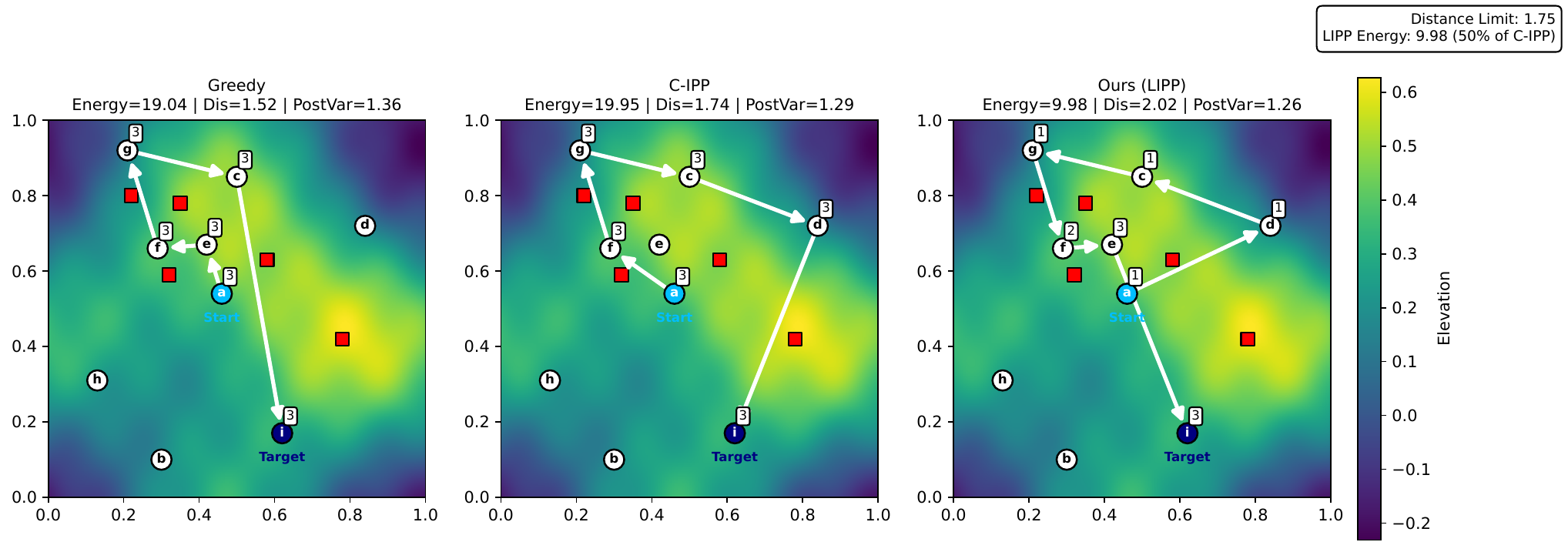}
    \caption{Comparison of sampling strategies on a synthetic scalar field, with the heatmap showing lunar surface elevation and white boxes indicating the number of samples collected at each vertex. \textbf{(left)} Greedy myopically selects the vertex with the greatest distance-normalized uncertainty reduction at each step, exhausting its budget before it can reach vertex~d. \textbf{(center)} \cipp\ plans globally and budgets its distance to include vertex~d, but samples the high-value region early, incurring high energy cost. \textbf{(right)} \lipp\ visits vertex~e in addition to all vertices \cipp\ visits, while ordering the route so that sample-intensive regions come later, achieving greater posterior variance reduction at half the energy.}

    \label{fig:th_heatmap_comparison}
\end{figure*}

Finally, although we have defined a theoretical bound, the path length can also be controlled explicitly by adding the linear constraint $\sum_{(u,v)\in\mathcal E} d_{uv}\,\chi_{uv} \le b$ for a user-specified budget $b$. This limits execution time while still optimizing load-dependent energy, without increasing the number of decision variables. We thus obtain theoretical execution-time bounds for arbitrary graphs, together with a simple mechanism for regulating execution time directly.

\begin{figure*}[t]
    \centering
    \subfloat[]{%
        \includegraphics[width=0.32\textwidth]{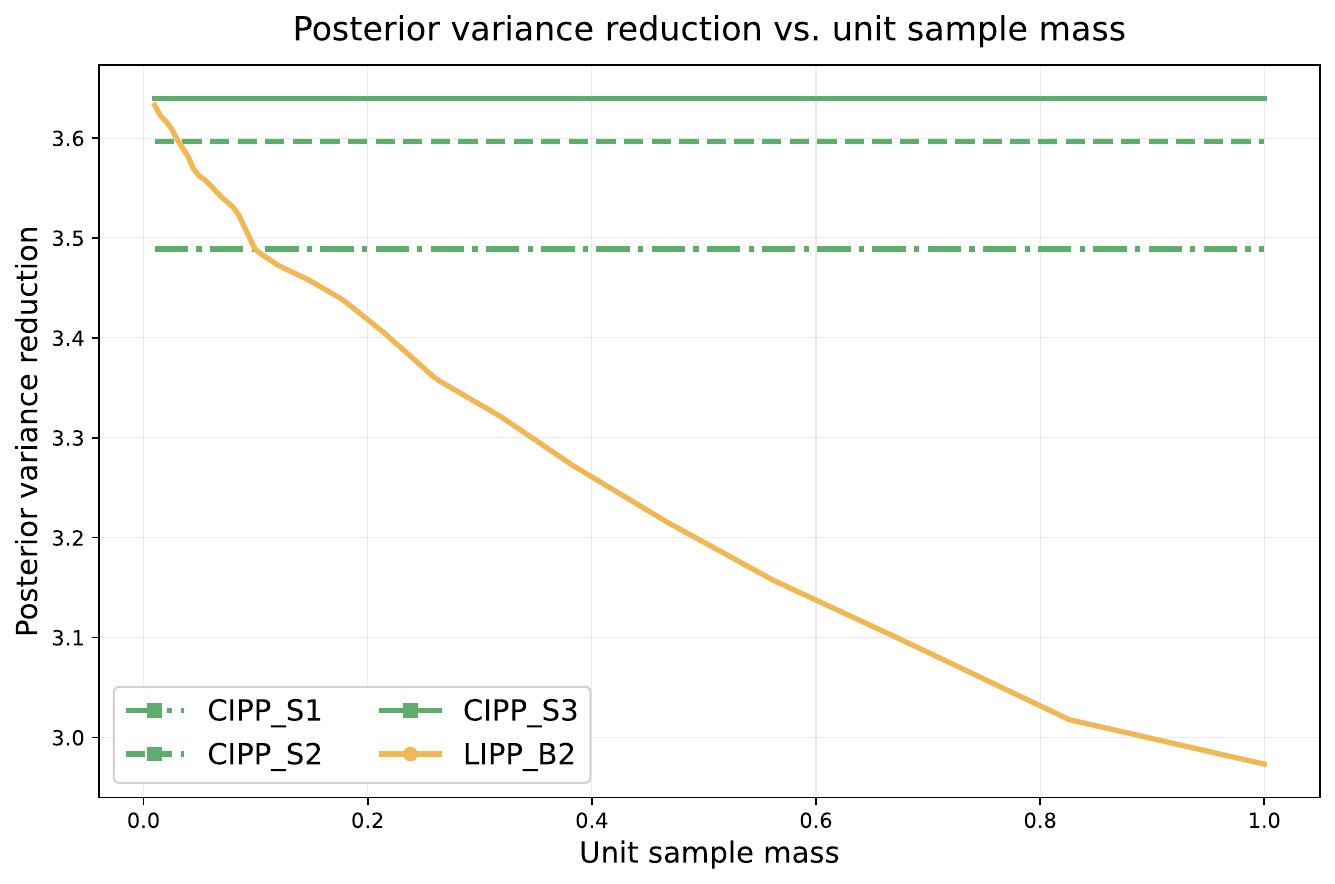}%
        \label{fig:ratio}
    }
    \hfill
    \subfloat[]{%
        \includegraphics[width=0.32\textwidth]{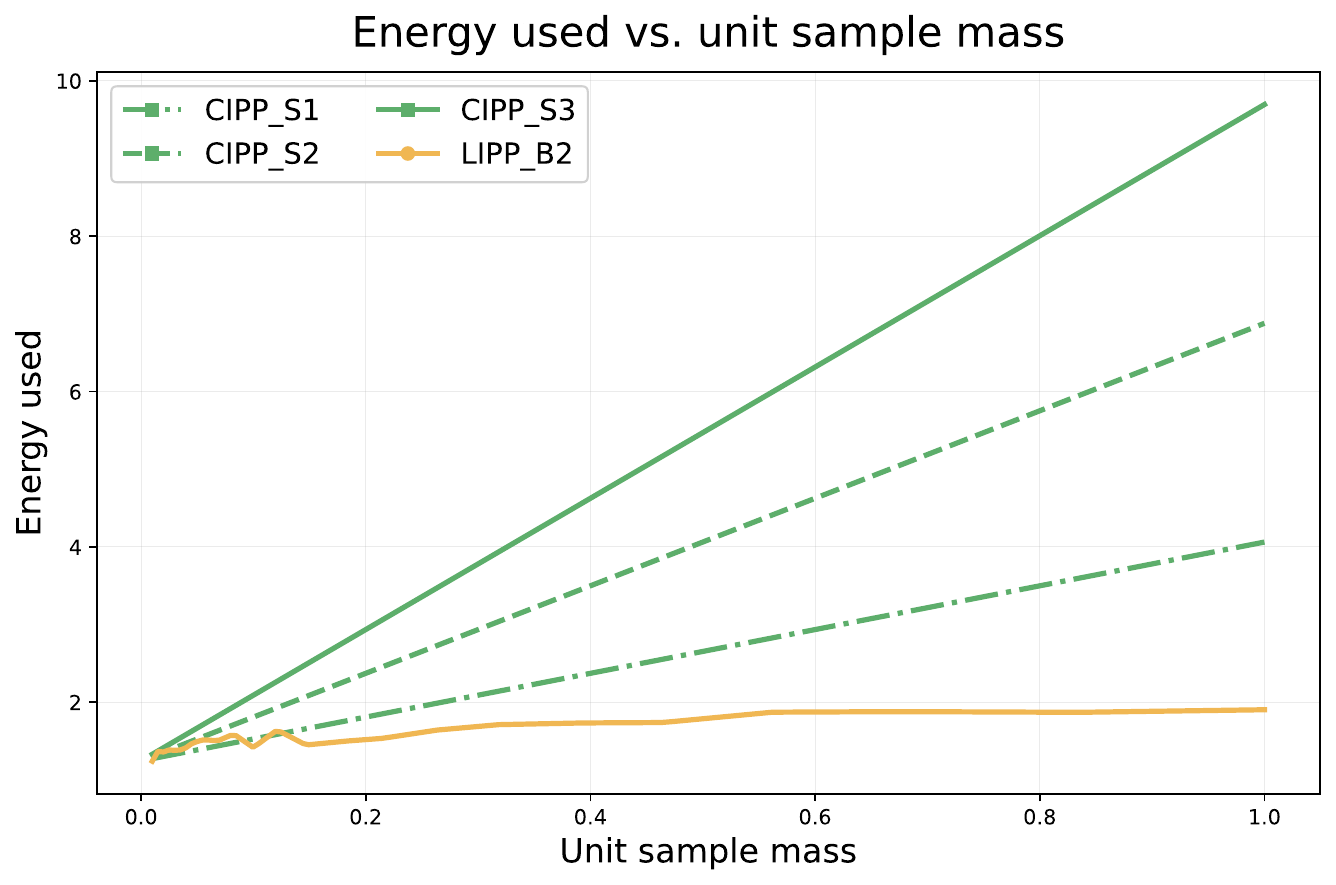}%
        \label{fig:energy_unit_mass}
    }
    \hfill
    \subfloat[]{%
        \includegraphics[width=0.32\textwidth]{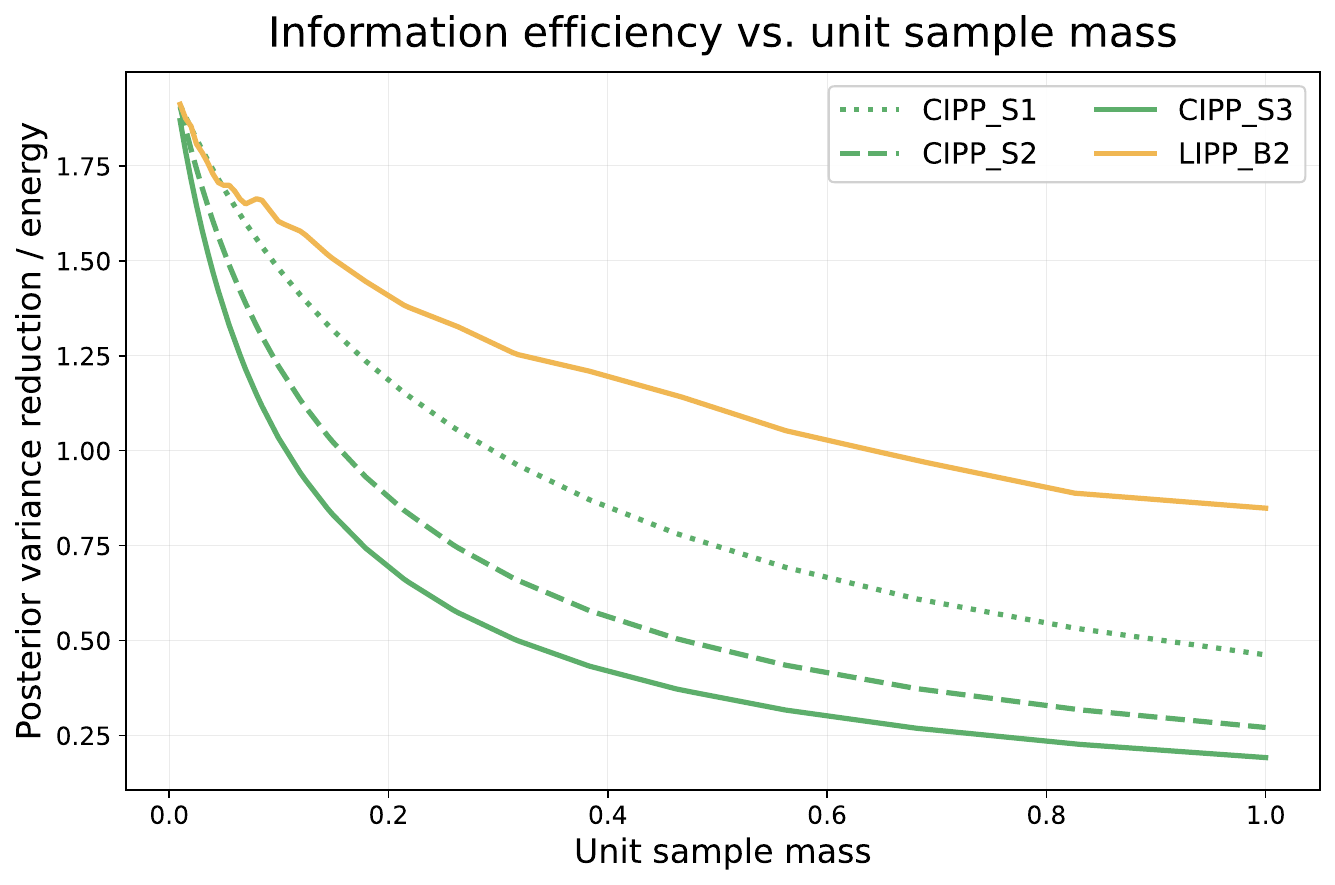}%
        \label{fig:efficiency_unit_mass}
    }
    \caption{Posterior variance reduction, total energy used, and posterior variance reduction per unit energy averaged over 2{,}000 randomly generated graphs with $R_0=1.0$ across different unit sample mass $\lambda$.
    \textbf{(a)} As the unit sample mass approaches zero, the energy constraint relaxes and \lipp\ converges to the same posterior variance reduction as \cipp.
    \textbf{(b)} As the unit sample mass increases, \lipp\ expends energy much more slowly than \cipp\ and is bounded by its energy budget B=2.
    \textbf{(c)} As the unit sample mass increases, \lipp\ achieves progressively greater reduction in posterior variance per unit energy than \cipp.}
    \label{fig:lambda}
\end{figure*}

\subsection{Computational Complexity}
Both formulations result in mixed-integer quadratic programs (MIQPs), but they differ in relaxation strength and solver behavior. In terms of nominal problem size, the \cipp formulation scales as $O(m n^2)$, whereas the \lipp formulation introduces additional sampling-indexed variables and bilinear terms, increasing the size to $O(m n^2 S_{\max}^2)$.

In practice, however, the computational gap is driven less by polynomial growth and more by differences in relaxation quality. In the classical formulation, routing and sampling variables are largely separable, leading to comparatively tighter continuous relaxations. In contrast, the \lipp formulation—through the bilinear coupling introduced in Constraint \eqref{eq:omega_bounds}—links load accumulation, sampling decisions, and traversal order. When integrality constraints are relaxed, these interactions permit fractional routing and load propagation, which can weaken the resulting lower bound after linearization (e.g., via McCormick envelopes) \cite{Vielma2015}. A weaker root relaxation forces the branch-and-bound solver to explore a larger portion of the search tree to certify optimality. The empirical solve-time trends in Fig.~\ref{fig:runtime} suggest this behavior is realized in practice as problem size grows. 

\section{Experimental Results}

In this section, we empirically validate four properties of \lipp\ against \cipp\ and a Greedy baseline over 2{,}000 randomly generated scenarios spanning various graph densities and sizes: (i) the added sampling-amount dimension changes the optimal vertex set, not merely visitation order; (ii) \lipp\ coincides with \cipp\ at $\lambda=0$ and outperforms it as $\lambda$ grows; (iii) it achieves comparable paths at lower energy under reasonable budgets; and (iv) its overhead stems mainly from weaker relaxations.

\subsection{Experiment Setup}

The experiment emulates a lunar rover mission collecting regolith samples for subsequent estimation of thorium (Th) surface concentration, which provides critical geochemical information such as crustal composition and evolution history~\cite{Lawrence1998LunarProspector}. Terrain elevation is incorporated by modeling traversal cost as a function of elevation difference between vertices, $d_{uv} = d_{\mathrm{euclid}} \cdot (1 + \alpha(\mathrm{height}_v - \mathrm{height}_u))$, where $\alpha$ is a constant scaling factor. The total energy to traverse an edge is the traversal cost multiplied by the robot's accumulated mass at the start of that edge. The elevation profile and Th concentration field are held fixed across all trials. The \cipp\ formulation follows Section~V-A, and the Greedy heuristic selects at each step the vertex that maximizes posterior variance reduction normalized by travel distance. For \cipp\ and Greedy, we evaluate uniform sample counts of 1, 2, and 3 per visited vertex, denoted by the suffixes ``\_S1'', ``\_S2'', and ``\_S3'', respectively. For \lipp, the suffix ``\_Bk'' indicates an energy budget of $k$ units.

\begin{figure*}[t]
    \centering
    \subfloat[]{%
        \includegraphics[width=0.32\textwidth]{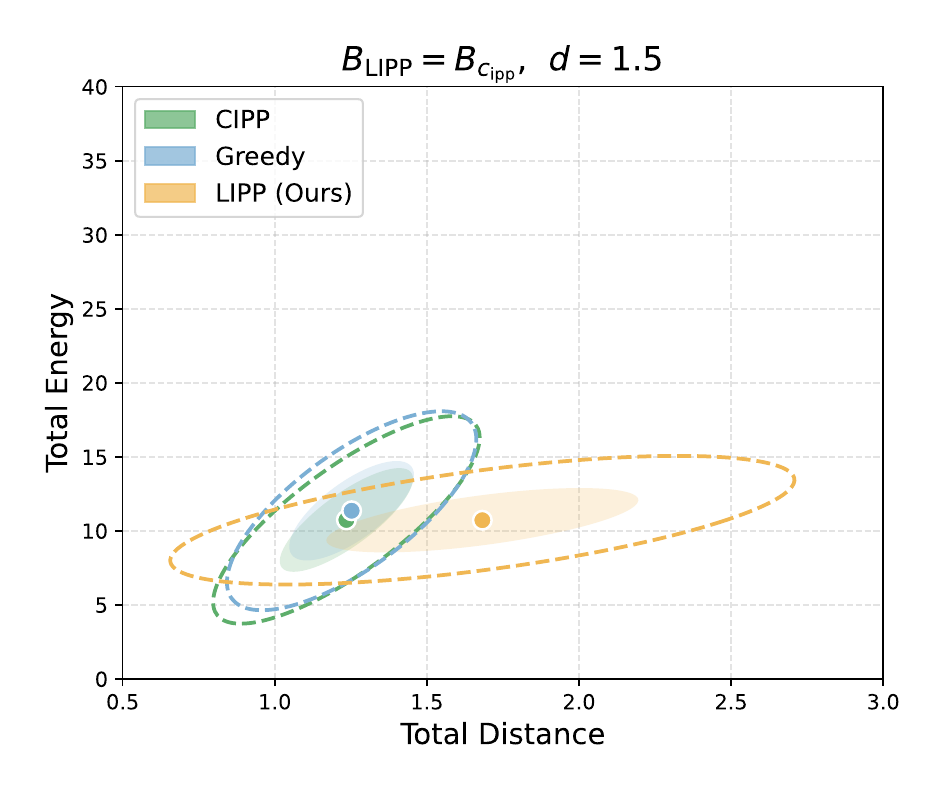}%
        \label{fig:lambda_distance_a}
    }
    \hfill
    \subfloat[]{%
        \includegraphics[width=0.32\textwidth]{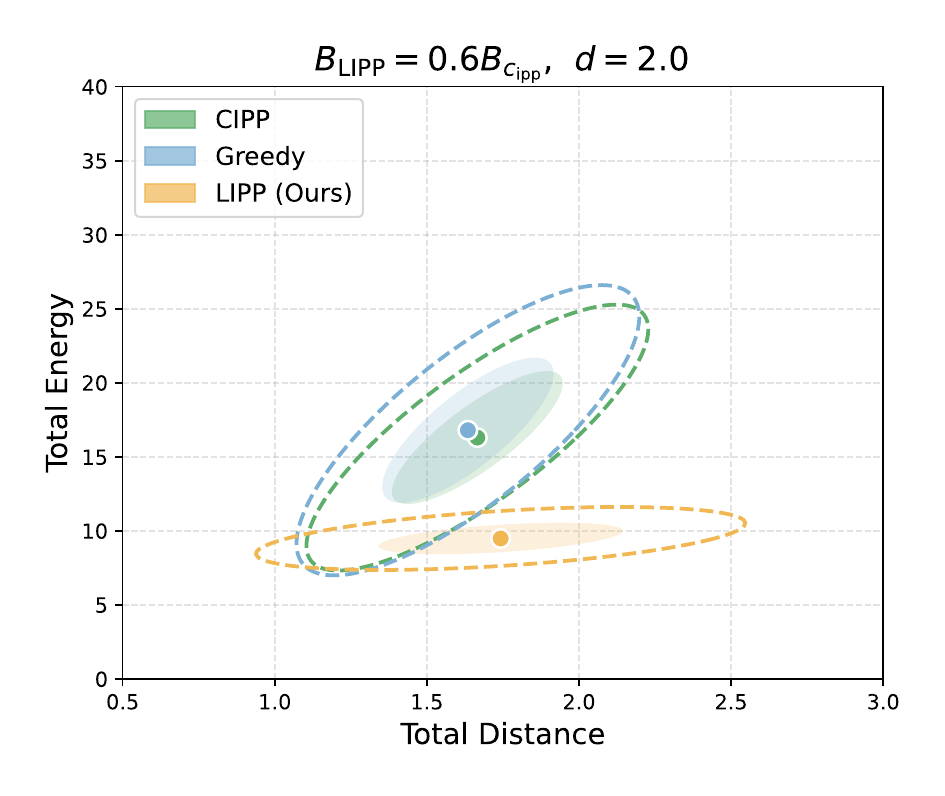}%
        \label{fig:lambda_distance_b}
    }
    \hfill
    \subfloat[]{%
        \includegraphics[width=0.32\textwidth]{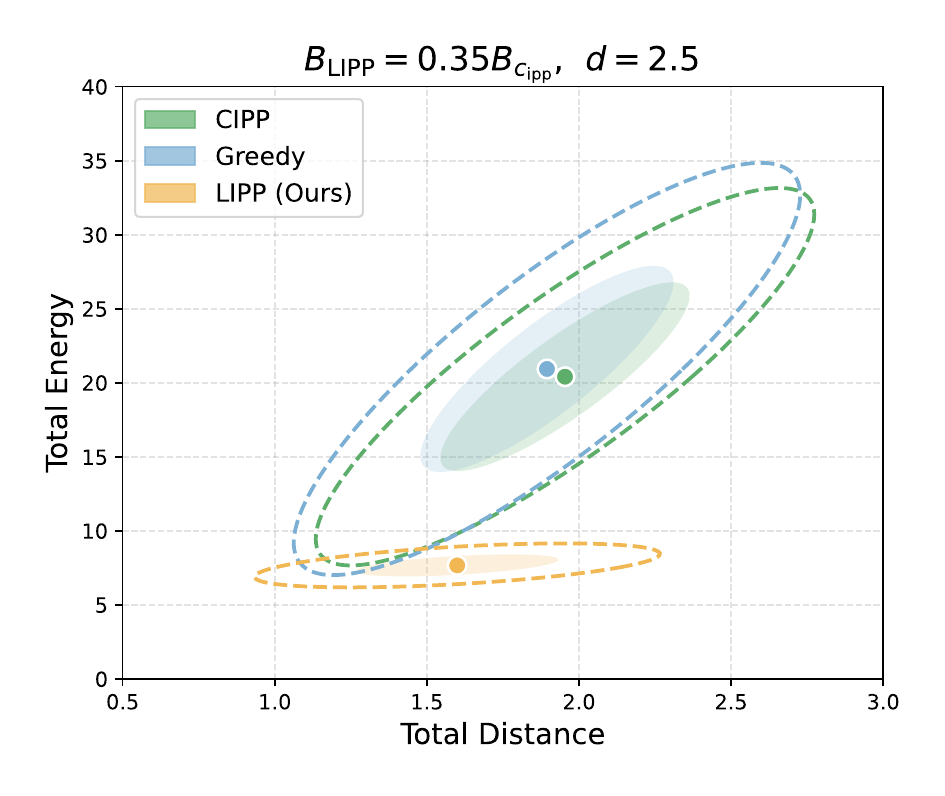}%
        \label{fig:lambda_distance_c}
    }
    \caption{Distance--energy trade-off of \lipp\ against \cipp\ and Greedy across three budget regimes, over 2{,}000 randomly generated graphs with $S_{\max}=3$. For each graph we run \cipp, record its path length and consumed energy $B_{\mathrm{CIPP}}$ (taking $S_{\max}$ samples per vertex), and allocate \lipp\ a fraction of $B_{\mathrm{CIPP}}$: \textbf{(a)} the full $B_{\mathrm{CIPP}}$, \textbf{(b)} $0.6\,B_{\mathrm{CIPP}}$, and \textbf{(c)} $0.35\,B_{\mathrm{CIPP}}$. To compare distance and energy fairly, we retain only instances where all three methods reach similar posterior variance on the same graph (within $\pm 0.1$), discarding cases where a method sacrificed accuracy. Given a generous budget (a), \lipp\ takes longer detours; as the budget tightens (b, c), it matches or beats \cipp\ in both distance and energy.}
    \label{fig:dist_energy}
\end{figure*}

\subsection{Qualitative Results}
To validate property (i), Fig.~\ref{fig:th_heatmap_comparison} compares the paths each method produces on the same instance. The Greedy algorithm recursively selects the vertex with the greatest distance-normalized posterior-variance reduction, yielding a myopic path that exhausts its budget before reaching vertex~d. \cipp\ plans over a longer horizon and, despite the tight distance budget, routes to include vertex~d; its distance constraint, however, prevents it from also sampling at vertex~e. \lipp\ defers heavy sampling to the end of the path: it collects a single unit at vertices~d, c, and~g, far from the test locations, then increases the sample count in the important region around vertices~f and~e, where the test locations are clustered. It achieves this while expending roughly half the energy of \cipp. Crucially, because sampling amount is an added decision dimension, \lipp\ does more than redistribute samples over a fixed vertex set---it selects a different set of vertices altogether, altering both sampling allocation and routing rather than merely permuting visitation order.

\subsection{Quantitative Results}
To validate property (ii), i.e., that \lipp\ relaxes to \cipp\ as discussed in Section~III-B, we set the energy budget of \lipp\ to $B = 2.0$, the distance budget of \cipp\ to $b = 2.0$, the robot mass to $R_0 = 1.0$, and sweep the unit sample mass $\lambda$ from $0$ to $1$. Fig.~\ref{fig:ratio} and \ref{fig:efficiency_unit_mass} show that as $\lambda \to 0$, the posterior variance reduction and efficiency of \lipp\ converge to those of \cipp\ with $S = S_{\max}$. This matches the theoretical equivalence in Section~III-B: when samples are weightless, \lipp\ naturally collects the maximum number at every visited vertex. The posterior variance reduction of \cipp, by contrast, stays flat as sample mass grows, showing that its path selection is entirely decoupled from the physical cost of carrying samples. This decoupling drives a steep rise in total energy consumption (Fig.~\ref{fig:energy_unit_mass}). \lipp\ instead explicitly accounts for the growing load through its joint optimization (Section~IV), adapting routing order and sampling amount to the energy budget. At $\lambda = 1.0$, Fig.~\ref{fig:efficiency_unit_mass} shows \lipp\ attaining roughly three times the posterior variance reduction per unit energy of the \cipp\ baseline under uniform sampling with $S_{\max} = 3$. This demonstrates that explicitly modeling load-dependent coupling yields substantial efficiency gains when the physical burden of carrying samples is non-negligible.

Regarding property (iii), like any informative path planning problem, \lipp's behavior depends heavily on the budget it is given. To examine how much path length \lipp\ sacrifices to save energy, we test three energy budgets, each defined relative to $B_{\mathrm{CIPP}}$, the energy \cipp\ consumes on the same graph: an excessive budget equal to the full $B_{\mathrm{CIPP}}$, a moderate budget of $0.6\,B_{\mathrm{CIPP}}$, and a tight budget of $0.35\,B_{\mathrm{CIPP}}$. Given the full $B_{\mathrm{CIPP}}$ (Fig.~\ref{fig:lambda_distance_a}), \lipp\ has enough energy to follow the same path as \cipp\ and collect the maximum number of samples at each vertex; instead, it tends to explore additional vertices rather than gather more information at a few, though never exceeding the $\frac{R_0 + \lambda S_{\max}(p_D - 1)}{R_0 + \lambda}$ distance bound of Section~V-A. As the budget tightens to $0.6\,B_{\mathrm{CIPP}}$ (Fig.~\ref{fig:lambda_distance_b}), this slack disappears: \lipp\ matches \cipp's travel distance while attaining comparable posterior variance at lower energy. Under the tight $0.35\,B_{\mathrm{CIPP}}$ budget (Fig.~\ref{fig:lambda_distance_c}), \lipp\ can afford neither heavy sampling nor every useful vertex, which constrains it to shorter paths than \cipp\ while still reaching the same posterior variance. Note that we retain only graphs on which all methods achieve similar posterior variance reduction; consequently, the $0.35\,B_{\mathrm{CIPP}}$ cases tend to be graphs where a short path already suffices, and the full-budget cases those where it does not. These results indicate that \lipp's longer paths are not an inherent limitation but a consequence of an overly generous budget; under a suitably tight budget, it matches or improves upon \cipp\ in both distance and energy.

\begin{figure}[h]
    \centering
    \subfloat[]{\includegraphics[width=0.48\columnwidth]{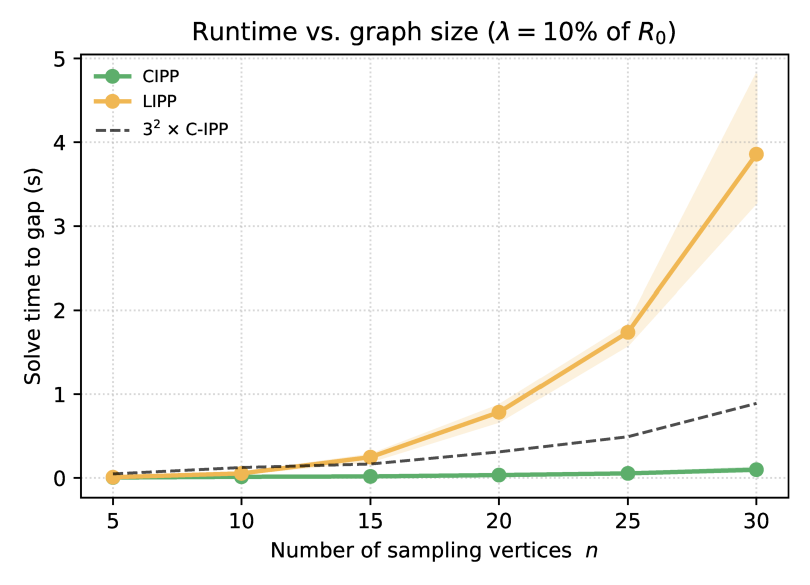}\label{fig:node_sweep}}%
    \hfill
    \subfloat[]{\includegraphics[width=0.48\columnwidth]{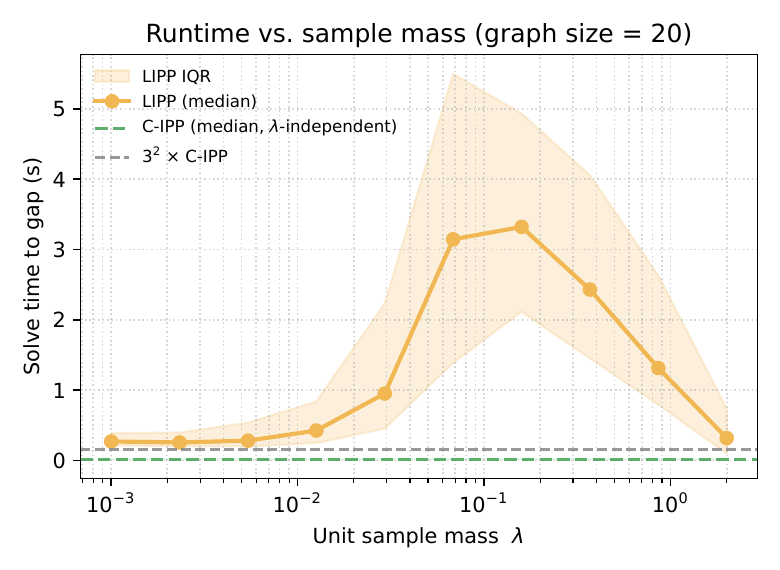}\label{fig:lambda_sweep}}%
    \caption{Runtime comparison of \cipp\ and \lipp, measured on 2{,}000 random graphs; solve time is the time for Gurobi to reach a relative optimality gap below 5\%. \textbf{(a)} Solve time versus graph size at fixed sample mass and density ${\approx}15\%$: \cipp\ stays tractable, whereas \lipp\ grows steeply and exceeds the $S_{\max}^2 = 9 \times$ \cipp\ reference. \textbf{(b)} Solve time versus unit sample mass $\lambda$ at fixed graph size of 20: at small $\lambda$, \lipp\ tracks the $S_{\max}^2 \times$ \cipp\ reference (the variable-count factor alone), rises above it as the load coupling strengthens, then falls once a heavy sample mass shrinks the feasible set.}
    \label{fig:runtime}
\end{figure}

Lastly, regarding property (iv), the primary limitation of \lipp\ is computational cost. As shown in Fig.~\ref{fig:node_sweep}, both \cipp\ and \lipp\ scale steeply with graph size, as expected for NP-hard mixed-integer programs, but \lipp\ grows markedly faster than \cipp. This overhead has the two sources identified in Section~V-B: the $S_{\max}^2$-fold increase in decision variables and, dominantly, a weaker continuous relaxation that enlarges the branch-and-bound search space. Fig.~\ref{fig:lambda_sweep} disentangles the two against the $S_{\max}^2 \times$ \cipp\ reference line, which captures the variable-count factor alone: when the coupling is negligible (small $\lambda$), \lipp\ tracks this reference, whereas at non-trivial sample mass it rises well above it, showing that the relaxation---not the variable count---drives the excess cost. Fig.~\ref{fig:lambda_sweep} further reveals a non-monotonic trend: once samples become heavy enough that the energy budget sharply restricts the feasible set, the search space shrinks and solve time falls again. Although planning is performed offline, scaling \lipp\ to the finer discretizations needed for real-world deployment will require more efficient exact and approximate solvers.

\section{Conclusion and Future Work}

In this work, we introduced \lipp (load-aware \ipp), motivated by a fundamental limitation of classical \cipp: when information acquisition physically alters the robot through cumulative sampling load, the traversal cost of future edges depends on past sampling decisions, making the problem inherently order-dependent — a property \cipp cannot model. We showed that \lipp strictly generalizes \cipp, recovering it exactly as sample unit mass $\lambda \to 0$, and derived a MIQP formulation that jointly optimizes routing, visitation order, and sampling allocation under an energy budget. We further established theoretical bounds on the path-length increase of \lipp relative to \cipp and validated these properties across 2{,}000 diverse mission scenarios, demonstrating that \lipp achieves comparable posterior variance at significantly lower energy cost as sample mass grows.

\lipp opens several directions for future work. First, the weaker LP relaxations introduced by load-dependent coupling motivate faster solution methods—such as tailored heuristics, tighter relaxations, or approximation schemes—to scale to larger environments. Second, \lipp lays the groundwork for collaborative, heterogeneous multi-robot informative path planning with physical sampling. For example, a mission might pair agile scouts capable of rapid sampling with slower carrier robots that transport large accumulated loads at lower relative energy cost. Such settings amplify the importance of explicitly modeling load-dependent mobility, and constitute a promising direction for future research.


\bibliographystyle{IEEEtran}
\bibliography{references}

\end{document}